\documentclass{article}
\usepackage{spconf,amsmath,graphicx,url, multirow}

\title{Analysis and Detection of Differences in Spoken User Behaviors between Autonomous and Wizard-of-Oz Systems}

\name{Mikey Elmers, Koji Inoue, Divesh Lala, Keiko Ochi, and Tatsuya Kawahara}
\address{Graduate School of Informatics\\
    Kyoto University, Japan\\
}

\begin{document}

\maketitle

\begin{abstract}
This study examined users' behavioral differences in a large corpus of Japanese human-robot interactions, comparing interactions between a tele-operated robot and an autonomous dialogue system.
We analyzed user spoken behaviors in both attentive listening and job interview dialogue scenarios.
Results revealed significant differences in metrics such as speech length, speaking rate, fillers, backchannels, disfluencies, and laughter between operator-controlled and autonomous conditions.
Furthermore, we developed predictive models to distinguish between operator and autonomous system conditions.
Our models demonstrated higher accuracy and precision compared to the baseline model, with several models also achieving a higher F1 score than the baseline.
\end{abstract}

\begin{keywords}
spoken dialogue system, human-robot interaction, corpus analysis, spoken behaviors

\end{keywords}

\section{Introduction}
\label{sec:introduction}
Conversational robots and spoken dialogue systems (SDSs) have become increasingly prevalent in various aspects of daily life.
Despite their growing presence, a large gap remains between conversational robots and human-like interaction.
Semi-autonomous systems offer a solution by providing a powerful combination of autonomous functionality with the capability to transfer control to human operators, when necessary.
These systems assist an operator in managing multiple remote robots, effectively handling issues that may arise \cite{kawahara2021robotics}.
However, effective evaluation of open dialogue systems is a complex process that requires automation, reproducibility, differentiation between systems, and explainability \cite{deriu2021survey}.
Identifying users' spoken behaviors that distinguish operator-controlled systems from autonomous systems is, therefore, crucial for improving the quality of semi-autonomous systems, cuing an operator for timely and appropriate intervention.
This methodology enables researchers to target their focus on areas where users' spoken behaviors differ greatly between the autonomous and operator-controlled systems.

\section{Methods}
\label{sec:methods}

This study analyzed a large corpus of human-robot interactions for both attentive listening and job interview scenarios.
We used an autonomous system and a Wizard-of-Oz (WoZ) style experiment, in which participants interacted with a robot under the impression that it was controlled autonomously.
In reality, it was controlled by a human operator.
Spoken user behaviors were evaluated to determine differences between the autonomous and WoZ systems.

\subsection{ERICA}
ERICA \cite{glas2016roman, inoue2016sigdial} is an autonomous android equipped with the capability to engage proficiently in diverse social contexts.
ERICA has a human-like appearance complemented by a synthesized voice, trained on the vocal characteristics of a Japanese voice actress.
ERICA generates a range of linguistic phenomena, including backchannels, fillers, and laughter.
Furthermore, ERICA produces a variety of visual cues, encompassing facial expressions, synchronized lip movements during speech, blinking, and nodding gestures.
ERICA showcases versatility by assuming distinctive roles, including attentive listening \cite{inoue2020sigdial}, conducting practice job interviews \cite{inoue2020icmi, kawai2022sigdial}, and serving as a laboratory guide \cite{inoue2021engagement}.
The primary objective of ERICA is to replicate human-like communication abilities across a broad range of scenarios.

\subsection{Attentive Listening}
Attentive listening involves the system actively engaging in speech perception, attentively listening to the subject's spoken communication.
Attentive listening with conversational robots offers notable benefits, particularly in offering companionship to elderly individuals \cite{inoue2020sigdial} and patients in psychiatric daycare settings \cite{ochi2023robotics}.
These systems demonstrate attentiveness by generating backchannels, asking follow-up questions, and offering empathetic responses.

During the sessions, subjects engaged in approximately 5 to 8-minute conversations with the robot ERICA, which was operated either autonomously or remotely piloted by an operator.
Conversation topics varied, with participants discussing subjects such as food, travel, or the challenges posed by the COVID-19 pandemic.
The attentive listening system used in this work was developed by \cite{inoue2020sigdial}.

\subsection{Job Interview}
In the job interview scenario, the robot ERICA assumes the role of the interviewer, conducting a practice interview with the interviewee—the user.
Job interviews are widely recognized as challenging and can evoke anxiety in candidates.
Interacting with an autonomous system offers prospective employees and students a valuable opportunity to practice and ready themselves for interviews, thereby mitigating performance-related anxieties.
Each session spanned approximately 9 minutes and was operated autonomously or remotely piloted by an operator.
The job interview system in this study was based on the work of \cite{inoue2020icmi}, integrating follow-up questions that required further elaboration from the interviewee.

\subsection{User Behavior}
This work builds upon the research conducted by \cite{inoue2024iwsds}, which similarly explored the analysis and evaluation of user behaviors in SDSs.
The current study evaluated several spoken user behaviors, including fillers, backchannels, disfluencies, and laughter.
While ERICA is capable of generating some of these phenomena as well, this study focused solely on the subjects' spoken behaviors.

Fillers play a crucial role in turn-taking and floor maintenance \cite{clark2002cognition}, and can serve as sociolinguistic identifiers \cite{fruehwald2016filled}.
Examples of fillers include ``ano'' and ``eto'' in Japanese, and ``uh'' and ``um'' in English.
In this work fillers are generated using the approach proposed by \cite{nakanishi2019iwsds}.

Backchannels convey attentiveness, interest, understanding, acknowledgement, agreement, and importantly, do not take the conversational floor \cite{ike2010backchannel}.
Active listening can be demonstrated through backchannels, elevating the listener to co-narrator status \cite{bavelas2000conarrator}, and adaptive backchanneling has been observed to encourage less talkative participants to engage more actively in the conversation \cite{cumbal2022acm}.
Examples of backchannels include ``hai'' and ``un'' in Japanese, and ``mhmm'' and ``uh-huh'' in English.
Backchannel frequency significantly differs between these two languages, with Japanese incorporating backchannels much more frequently than American English \cite{maynard1986linguistics}.
Previous research has explored the generation of verbal backchannels in both dyadic conversations \cite{lala2017sigdial} and group settings \cite{lala2022hai}.
This study focused on verbal backchannels, as opposed to nonverbal backchannels like nodding.

Disfluencies include a range of phenomena, such as lengthening, truncation, repair, or word fragmentation.
These phenomena can have both positive and negative effects for listeners.
For example, \cite{corley2007cognition} found that material preceded by a disfluency is more likely to be remembered, and \cite{macgregor2010neuropsychologia} found that disfluencies can provide additional processing time to listeners through temporal delay.
However, disfluencies can also carry negative implications, potentially leading listeners to perceive the speaker as less prepared or less knowledgeable.

Laughter functions as a tool in fostering social relationships in both human-human and human-robot contexts.
Laughter is pivotal for SDSs, contributing to emotional and affective engagement, natural language understanding, and pragmatic reasoning \cite{mazzocconi2022transactions}.
Additionally, mutual laughter has been linked to positive outcomes during job interviews, including increased likelihood of job offers for interviewees \cite{adelsward1989}, and the establishment of rapport \cite{brosy2021psychology}.
ERICA generates laughter using the approach outlined by \cite{inoue2022frontiers}.
 
\subsection{Experiment Setup}
We used inter-pausal units (IPU) as the transcription unit, with speech segments divided into distinct IPUs whenever a pause of 200 ms or longer was detected.
Furthermore, the transcription data was manually annotated to include additional linguistic information, including backchannels, fillers, disfluencies, and laughter.
Text tokenization into characters was performed using MeCab\footnote{\url{https://taku910.github.io/mecab/}}.
Length and speaking rate were measured using characters rather than words, as it offers a more stable metric.
Length was quantified as the number of tokenized characters within each IPU, and speaking rate was calculated by dividing the length of the IPU by its duration, measured in seconds.

\begin{figure}[t]
  \includegraphics[width=\columnwidth]{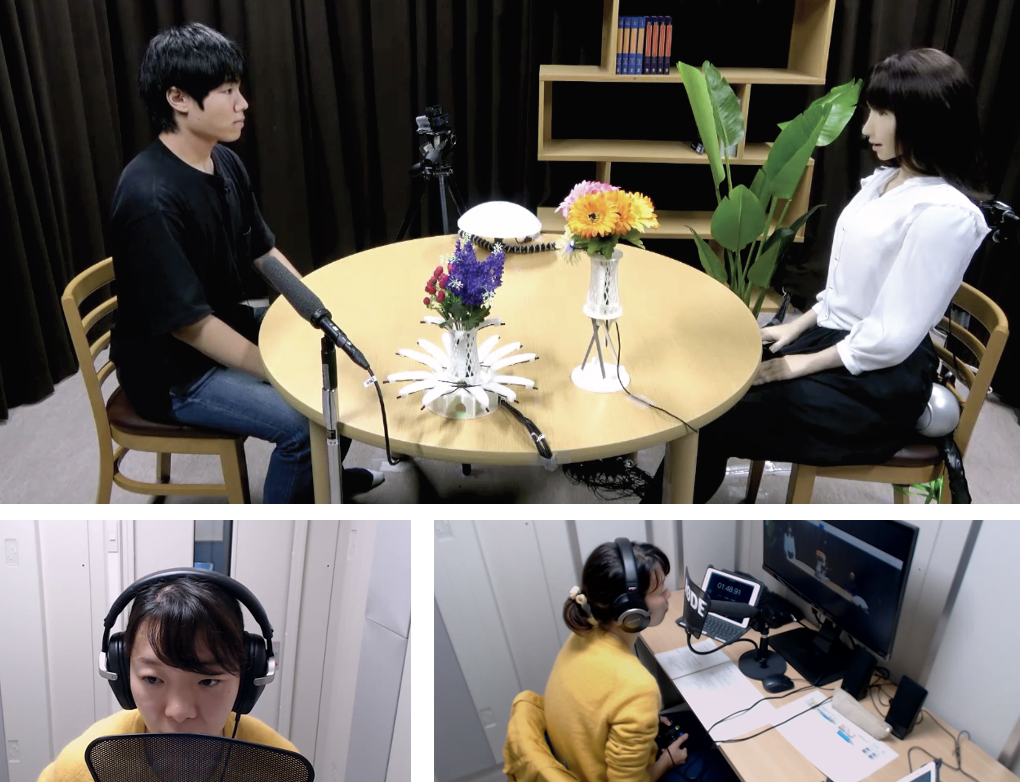}
  \caption{Illustration of experimental setup. The top frame is a side profile view of a subject (left side) and ERICA (right side). The bottom-left and bottom-right frames depict the operator's activity during the WoZ condition.}
  \label{fig:experiment_setup}
\end{figure}

This study investigated attentive listening and job interview scenarios for both WoZ and autonomous conditions.
During the WoZ condition, operators spoke into a microphone, and the audio was played through ERICA's speaker.
When users interacted with ERICA, a wide range of audio and video data was captured.
All interactions were conducted in Japanese.
Figure~\ref{fig:experiment_setup} show an example of the experimental setup.

\section{Results}
\label{sec:results}

\begin{table}
  \centering
  \small
  \begin{tabular}{lcccc}
    \hline
    \multicolumn{4}{c}{\textbf{Attentive Listening}} \\
    \hline
                  & \textbf{Auto} & \textbf{WoZ} & \textbf{p-value}\\
    \hline
    Length        & 9.91 (8.95)   & 10.63 (9.75) & $<$0.001\\
    Sp. Rate      & 6.04 (2.27)   & 6.56 (2.47)  & $<$0.001\\
    Fps           & 0.26 (0.67)   & 0.32 (0.88)  & $<$0.001\\
    Bps           & 0.34 (1.09)   & 0.42 (1.29)  & $<$0.001\\
    Dps           & 0.11 (0.67)   & 0.10 (0.79)  & $<$0.001\\
    Lps           & 0.04 (0.32)   & 0.06 (0.41)  & $<$0.001\\
    Filler        & 26.88\%       & 30.03\%      & $<$0.001\\
    Backchannel   & 10.69\%       & 11.77\%      & $<$0.001\\
    Disfluency    & 8.12\%        & 6.69\%       & $<$0.001\\
    Laugh         & 2.40\%        & 4.00\%       & $<$0.001\\
    \hline
    \multicolumn{4}{c}{\textbf{Job Interview}} \\
    \hline
                  & \textbf{Auto} & \textbf{WoZ} & \textbf{p-value}\\
    \hline
    Length        & 14.00 (13.33)  & 11.46 (11.27)& $<$0.001\\
    Sp. Rate      & 7.29 (2.50)   & 7.77 (2.91)  & $<$0.001\\
    Fps           & 0.48 (0.95)   & 0.46 (1.56)  & $<$0.001\\
    Bps           & 0.39 (1.27)   & 0.87 (2.03)  & $<$0.001\\
    Dps           & 0.07 (0.63)   & 0.09 (0.88)  & $>$0.05\\
    Lps           & 0.01 (0.13)   & 0.03 (0.29)  & $<$0.001\\
    Filler        & 46.1\%        & 30.1\%       & $<$0.001\\
    Backchannel   & 9.51\%        & 17.10\%      & $<$0.001\\
    Disfluency    & 6.49\%        & 5.80\%       & $>$0.05\\
    Laugh         & 0.49\%        & 2.51\%       & $<$0.001\\
    \hline
  \end{tabular}
  \caption{Descriptive and inferential statistics.} 
  \label{tab:combined_stats}
\end{table}

For the attentive listening scenario, there were 109 sessions for the WoZ condition and 100 sessions for the autonomous condition\footnote{All data and code for the results section can be accessed at \url{https://github.com/MikeyElmers/paper_ococosda24}}.
This resulted in 23,662 WoZ IPUs and 16,902 autonomous IPUs for the attentive listening scenario.
For the job interview scenario, there were 29 sessions for the WoZ condition and 44 sessions for the autonomous condition. 
This resulted in 4,414 WoZ IPUs and 4,533 autonomous IPUs for the job interview scenario.
These IPU counts refer only to the subjects' speech and exclude IPUs from the WoZ operator or autonomous system.

Descriptive and inferential statistics are presented in Table~\ref{tab:combined_stats} for the attentive listening and job interview scenarios for the autonomous (Auto) and WoZ conditions.
Mean values and standard deviations (in parentheses) are reported for IPU length, speaking rate, fillers per second (fps), backchannels per second (bps), disfluencies per second (dps), and laughs per second (lps).
The percentage of IPUs containing fillers, backchannels, disfluencies, and laughs is also provided. 
The data exhibited violations of normality, as observed through visual inspection and confirmed by the Shapiro-Wilk test, along with violations of the homogeneity of variances, as indicated by Levene's test.
The non-parametric Wilcoxon rank sum test was used for analyzing IPU length, speaking rate, fps, bps, dps, and lps.
The Chi-squared test was utilized for analyzing count data for fillers, backchannels, disfluencies, and laughs.
Since multiple comparisons were conducted, we applied a correction to control for the increased risk of Type I errors (false positives).
With an original $\alpha$ level of 0.05 and 10 comparisons, we used the Bonferroni correction, which adjusted the significance level to $\alpha$ = 0.005.
All features were statistically significant for the attentive listening scenario (\textit{p} $<$ 0.001).
All features were statistically significant for the job interview scenario (\textit{p} $<$ 0.001) except the percentage of IPUs containing disfluencies and the frequency of disfluencies per second.

\subsection{Attentive Listening}
Our analysis revealed that subjects exhibit longer IPU lengths and speak at a faster rate (with greater variability) in the WoZ condition compared to the autonomous condition.
This observed increase in the attentive listening scenario suggests that the WoZ condition effectively encourages subjects to speak more frequently and at a faster pace.

For fillers, we observed an increase in usage for the WoZ condition in the attentive listening scenario.
This phenomenon could stem from the self-driven nature of the attentive listening task.
We also noted an increase in both the mean and standard deviation of fillers per second for the WoZ condition in the attentive listening scenario.
This suggests that the WoZ operator is adept at displaying natural behaviors and eliciting a higher usage of fillers per second from the subjects.

We found an increase in backchannel presence and a higher rate of backchannels per second for the WoZ condition for attentive listening.
This indicates that subjects exhibit more active listening behaviors, even when engaging in predominantly one-sided conversations with the WoZ operator, compared to interactions with the robot.

We noted a reduction in disfluency usage and the frequency of disfluencies during the attentive listening scenario.
This decrease could be attributed to the WoZ operator being less distracting and eliciting fewer disfluency errors compared to the robot.

We observed a significant increase in the subject's usage of laughter, accompanied by an increase in both the mean and standard deviation for the frequency of laughter, in the WoZ condition during the attentive listening scenario.
This suggests that the WoZ operators are more capable of creating an environment for eliciting laughter in these situations.

\subsection{Job Interview}
Our analysis indicated that subjects use shorter IPU lengths (with less variability) in the WoZ condition compared to the autonomous condition.
An increase in speaking rate was observed in this scenario, which could potentially be attributed to nervousness caused by the job interview scenario.

We noted a significant decrease in filler usage for the WoZ condition in the job interview scenario.
Given the demanding nature of job interviews, it is plausible that subjects would strive to minimize their use of fillers, with the WoZ operator better simulating real-world job interview conditions and thereby prompting subjects to employ fewer fillers.
We also found a minor decrease in the average fillers per second, coupled with a substantial increase in standard deviation for the WoZ condition.
The large standard deviation suggests considerable variability among subjects in their utilization of fillers when interacting with the WoZ operator, potentially reflecting varying degrees of nervousness among individuals.

We found an increase in the occurrence of backchannels and an elevated rate of backchannels per second for the WoZ condition in the job interview scenario.
This substantial increase in backchannel usage suggests that participants may exhibit greater interest when conversing with the WoZ operator compared to the autonomous system.

We observed an overall decrease in disfluency usage during the job interview scenario for the WoZ condition.
This reduction could be attributed to the WoZ operator being less distracting, which may have led the subject to produce fewer disfluencies compared to when interacting with the autonomous system.
However, it is noteworthy that disfluencies and disfluencies per second were the only features in the job interview scenario that did not exhibit significant differences between the conditions.

We found a large increase in the subject's usage of laughter, and in the mean and standard deviation for the frequency of laughter, in the WoZ condition for the job interview scenario.
This suggests that the WoZ operator can cultivate a more humorous atmosphere even within the context of a job interview.
Such elements are significant in an interview context, as laughter and humor can serve as effective icebreakers, alleviate tension, and build rapport.

\subsection{Modeling}
We also conducted predictive modeling on both the attentive listening and job interview data.
The goal was to predict whether the subject was speaking to an operator (WoZ) or the autonomous system via information gathered from the subject for each IPU.
The predictive features were the same for all models and included the following: number of backchannels, backchannels per second, number of laughs, laughs per second, number of fillers, fillers per second, number of disfluencies, disfluencies per second, IPU length, speaking rate, and mean and standard deviation for both f0 and power.
All measurements were calculated for each IPU.
Separate training-test splits were generated for both the attentive listening and job interview datasets.
Random sampling was used to generate an 80/20 training-test split, with a preset seed value to ensure reproducibility.
To prevent speaker identification from influencing the results, the subject's IPUs were allocated exclusively to either the training split or the test split.
The default hyperparameters were used for all modeling, and no further tuning was conducted.

\begin{table}
  \centering
  \small
  \begin{tabular}{lcccc}
    \hline
    \multicolumn{5}{c}{\textbf{Attentive Listening}} \\
    \hline
    \textbf{Model} & \textbf{Acc} & \textbf{Prec} & \textbf{Recall} & \textbf{F1} \\
    \hline
    baseline        & 0.64         & 0.64          & 1.00            & 0.78        \\
    lg              & 0.66         & 0.69          & 0.86            & 0.76        \\
    svm             & 0.71         & 0.73          & 0.87            & 0.79        \\
    rf              & 0.70         & 0.74          & 0.81            & 0.77        \\
    \hline
    \multicolumn{5}{c}{\textbf{Job Interview}} \\
    \hline
    \textbf{Model} & \textbf{Acc} & \textbf{Prec} & \textbf{Recall} & \textbf{F1} \\
    \hline
    baseline        & 0.51         & 0.49          & 1.00            & 0.66        \\
    lg              & 0.55         & 0.54          & 0.53            & 0.54        \\
    svm             & 0.67         & 0.66          & 0.68            & 0.67        \\
    rf              & 0.69         & 0.69          & 0.67            & 0.68        \\
    \hline
  \end{tabular}
  \caption{Model evaluation metrics.}
  \label{tab:combined_eval_metrics}
\end{table}

Evaluation metrics for all models can be found in Table~\ref{tab:combined_eval_metrics} for the attentive listening and job interview scenarios.
We evaluated several models: a baseline model, a logistic regression (lg) model, a support vector machine (svm) model with a Gaussian radial basis function kernel, and a random forest (rf) model.
Our metrics included accuracy (acc), precision (prec), recall, and F1 score.
The baseline model predicted the majority class for all instances in the test data (i.e., a recall score of 1).
For both the attentive listening and the job interview scenarios, all models performed better than the baseline in terms of accuracy and precision.
For F1 score, the SVM model outperformed the other models for the attentive listening scenario, while the random forest model outperformed the other models in the job interview scenario.

To investigate the impact of each predictor, we conducted permutation-based variable importance analysis on the random forest models for both the attentive listening and job interview datasets.
This analysis quantified the significance of each feature by computing the change in model performance (i.e., accuracy) by randomly permuting the feature in question while keeping the other features constant.
Lower importance scores indicate a smaller effect on model performance, whereas higher scores denote a larger impact.
The scale of importance scores can vary depending on factors like the dataset, model, and evaluation metric employed.
Thus, it is important to interpret these scores relative to each other rather than focusing solely on their absolute values.
We conducted this analysis separately for each dataset, yielding permutation feature importance scores for each predictor variable.
Table~\ref{tab:vip_att_ji} shows the rankings of feature importance.

In both the attentive listening and job interview datasets, acoustic features emerged as the primary drivers for the model's predictive performance.
We also evaluated an acoustic-only model and linguistic-only model.
The acoustic-only model performed slightly worse than the combined acoustic and linguistic features model, while the linguistic-only model performed significantly worse.
These findings collectively suggest that, although linguistic features are beneficial, they are insufficient on their own for accurately classifying whether the subject is interacting with an operator or an autonomous system at the IPU level.

\begin{table}
  \centering
  \small
  \begin{tabular}{lcc}
    \hline
    \textbf{Feature}  & \textbf{Attentive Listening} & \textbf{Job Interview} \\
    \hline
    Mean Power        & 0.28                & 0.21 \\
    Sp. Rate          & 0.24                & 0.05 \\
    Mean f0           & 0.21                & 0.22 \\
    SD f0             & 0.17                & 0.09 \\
    SD Power          & 0.14                & 0.08 \\
    Length            & 0.13                & 0.05 \\
    Fps               & 0.07                & 0.04 \\
    Bps               & 0.02                & 0.02 \\
    Filler Count      & 0.02                & 0.02 \\
    Dps               & 0.01                & 0.01 \\
    Backchannel Count & 0.01                & 0.01 \\
    Lps               & 0.00                & 0.00 \\
    Disfluency Count  & 0.00                & 0.00 \\
    Laugh Count       & 0.00                & 0.00 \\
    \hline
  \end{tabular}
  \caption{Variable importance for random forest model.}
  \label{tab:vip_att_ji}
\end{table}

\section{Discussion}
\label{sec:discussion}
This study analyzed a large corpus of spoken interactions with the robot ERICA.
We observed that the WoZ condition prompted users to modify their spoken behavior.
Additionally, we found disparities in the features between the attentive listening and job interview scenarios.
These discrepancies may arise from the distinct nature of each scenario; while the attentive listening scenario entails lower stress and self-directed engagement, the job interview scenario typically involves higher stress levels, potentially leading to nervousness among the subjects.

Data collection took place over several years, which resulted in certain analyses facing challenges stemming from evolving circumstances.
Furthermore, the diversity of topics discussed ranged from subjects like food and travel to emotionally charged topics such as hardships during the COVID-19 pandemic.
The dataset lacked sufficient granularity to assess potential age and gender-related differences.
Since the modeling focused solely on IPUs, only the immediate context was evaluated, ignoring the full history of the dialogue.
Consequently, we cannot elaborate on user sensitivity to previous turns or the influence of the interaction as a whole.
Future research should evaluate additional units of dialogue to address these issues.
While the specific features examined in this study may not be generalizable to other domains or robots, we hope that this research will inspire further exploration into the data and evaluation of the underlying mechanisms behind observed phenomena.
Future research should also extend beyond spoken features to encompass visual cues and explore multi-modal interactions with SDSs and robots.
This holistic approach will contribute to a more comprehensive understanding of human-robot interaction and facilitate the development of more effective evaluation methodologies for SDSs.

\section{Conclusion}
\label{sec:conclusion}

This study evaluated subject interactions with the robot ERICA in both attentive listening and job interview scenarios.
Our findings indicated that subjects exhibited distinct speech patterns when interacting with the WoZ robot compared to the autonomous system.
Subsequently, we investigated whether the condition could be predicted using spoken user behaviors as features.
Our predictive models surpassed the performance of the majority class baseline for accuracy and precision, with several models also surpassing the F1 score of the baseline.

\bibliographystyle{IEEEbib}
\bibliography{refs}

\end{document}